\documentclass[runningheads]{llncs}
\usepackage[T1]{fontenc}
\usepackage{graphicx}
\usepackage{hyperref}
\usepackage{subcaption}
\usepackage{xcolor}

\begin{document}
\title{MO-IOHinspector: Anytime Benchmarking of Multi-Objective Algorithms using IOHprofiler}
\titlerunning{MO-IOHinspector}
%
\author{
Diederick Vermetten\inst{1}, 
Jeroen Rook\inst{2},
Oliver L. Preu\ss\inst{3},
Jacob de Nobel\inst{1},
Carola Doerr\inst{4},
Manuel L\'opez-Iba\~nez\inst{5},
Heike Trautmann\inst{2,3},
Thomas B\"ack\inst{1}
}
\institute{
LIACS, Leiden University, Leiden, The Netherlands
\and 
University of Twente, Enschede, The Netherlands
\and
Paderborn University, Paderborn, Germany
\and
LIP6, CNRS, Sorbonne Universit\'e, Paris, France
\and 
Alliance Manchester Business School, University of Manchester, Manchester, United Kingdom
}
\authorrunning{D. Vermetten et al.}

\maketitle              
\begin{abstract}

Benchmarking is one of the key ways in which we can gain insight into the strengths and weaknesses of optimization algorithms. In sampling-based optimization, considering the anytime behavior of an algorithm can provide valuable insights for further developments. In the context of multi-objective optimization, this anytime perspective is not as widely adopted as in the single-objective context. In this paper, we propose a new software tool which uses principles from unbounded archiving as a logging structure. This leads to a clearer separation between experimental design and subsequent analysis decisions. We integrate this approach as a new Python module into the IOHprofiler framework and demonstrate the benefits of this approach by showcasing the ability to change indicators, aggregations, and ranking procedures during the analysis pipeline.

\keywords{Benchmarking Framework  \and Multi-Objective Optimization \and  Anytime Performance \and Robust Ranking \and Visualization}
\end{abstract}

\section{Introduction}

Benchmarking is a core aspect of the process of developing new optimization algorithms~\cite{BarDoeBer2020benchmarking} as it serves as a way to provide comparisons of a new algorithm to the state of the art, providing insight into its relative strengths and weaknesses. In this spirit, benchmarking is more than a competitive race for more performance but serves our \emph{understanding} of algorithm behavior. Robust benchmarking of existing algorithms has the potential to provide a wealth of knowledge about the status of algorithmic development for specific types of problems and drive new development of algorithmic ideas to fill the gaps. 

To extract as much information from a given benchmark setup as possible, we should not be limited to looking only at the final solution(s) returned by an algorithm~\cite{HanAugBroTus2022anytime}. By looking at the full trajectory of an algorithm, we can identify regions of complementarity not just on different problems but at different stages of the optimization process. Looking at the \emph{anytime performance} of an algorithm can thus showcase the potential, e.g., for dynamically switching between algorithms~\cite{DBLP:conf/ppsn/KostovskaJVNWED22}, or provide a more detailed understanding of the relation between the evaluation budget and the relative rankings of different algorithms~\cite{HanAugBroTus2022anytime}. 

While this view on anytime performance has been widely adopted in single-objective optimization (SOO)~\cite{HanAugMer2020coco}, it faces several challenging design decisions to translate to the multi-objective optimization (MOO) setting. This is mainly due to the additional degree of freedom imposed by having not a single objective value but a set of optimal trade-off solutions, leading to differences, e.g., between which performance indicator is favored for a particular study. While for a given indicator value an anytime performance can be logged (e.g. the hypervolume indicator in bi-objective BBOB~\cite{BroTusTusWag2016biobj}), this approach imposes an analysis decision on the experimental setup. 

However, there is a great variety of unary set-based performance indicators~\cite{AuBigCar2021perf}, ranging from Pareto-compliant indicators such as hypervolume~\cite{ZitThiLauFon2003:tec} and R2 (under the condition of a continuous, uniform distribution of (Tchebycheff) utility functions \cite{SchKersch24}) to indicators which focus on decision space diversity~\cite{Preuss24,GRIMME2021}. Pareto compliance of an indicator means that the indicator respects the Pareto dominance relation. Specifically, if one solution dominates another, the indicator assigns a strictly better value to the dominating solution.  Indicators focusing on decision space diversity are more targeted towards multi-modal problems than the former. As such, some indicators might be more suited for specific research questions than others, and any imposed indicator choice limits how the resulting data can be re-used for new types of comparison.  

In this paper, we argue that the \textbf{later choice of analysis methodology 
need not influence design decisions} related to experimental setup, or vice versa. Instead, we show the benefits of using a data-logging approach based on the idea of unbounded archives~\cite{FieEveSing2003tec}. This not only supports the ability to easily consider anytime performance but also provides the freedom to decide about the indicator independently from the experiment design. While this requires a tradeoff regarding the amount of data stored, this data can subsequently be shared and re-used for various other analyses by the same or different groups of researchers. 

To simplify this type of logging, subsequent data analysis, and visualization, we introduce \textbf{MO-IOHinspector}, which builds upon the IOHprofiler framework~\cite{IOHanalyzer,IOHexperimenter2024}. While initially developed as an extension of the COCO platform~\cite{HanAugMer2020coco} for SOO, we now present an extension of the modular benchmarking toolkit to the multi-objective domain. With this new module, we provide an integration with the popular PyMOO library~\cite{DBLP:journals/access/BlankD20}, which we use to showcase the available analysis methods by running a benchmark study on several well-known multi-objective evolutionary optimization algorithms (MOEAs). Among the included analysis methods, we also incorporate the recently proposed robust ranking methodology~\cite{DBLP:conf/gecco/RookHT24}, 
aimed at creating a reliable and meaningful ranking that takes into account the variability in performance across different problem instances and simultaneously checks for statistical significance of performance differences. This usually results in groups of algorithms that are internally tied regarding performance but have different ranks across groups.

\section{Background and Related Work}

\subsection{Anytime Performance and Archiving}

When faced with an optimization problem, the amount of evaluation budget available is one of the critical aspects determining what algorithm or algorithm configuration would be most appropriate. As such, clearly understanding the relation between the number of function evaluations and the relative ranking of a set of algorithms is often an essential goal of benchmarking studies. Rather than looking only at the performance at the end of a fixed budget, looking at the full search trajectory provides some insights into this budget-dependence~\cite{HanAugBroTus2022anytime}. This is often viewed as \textit{anytime performance analysis} and has been widely used in benchmarking frameworks such as COCO~\cite{HanAugMer2020coco} and IOHprofiler~\cite{IOHprofiler}. 

In SOO, the common practice is to look at the evolution of the best-so-far objective value over time, which ensures a monotonic performance measure. This anytime analysis has led to the creation of powerful algorithm selection wizards, which are budget-dependent~\cite{MeuRakWon2022ngopt}. Additionally, comparing areas of the search trajectories where different algorithms perform well has led to studies into the potential of dynamically switching between optimizers~\cite{DBLP:conf/ppsn/KostovskaJVNWED22}.

The anytime analysis of multi-objective optimizers is more challenging~\cite{JesusAnytimeBiobj,JesusAnytimeIndicatorBB}, not only because of the lack of consensus on how to measure the quality of a population of solutions but also due to the difficulty of ensuring a monotonic increment of quality over time unless the population update (or archiving strategy) satisfies certain prerequisites, which is not the case for many multi-objective optimizers~\cite{LiLopYao2023archiving,SchHer2021archiving}. As a result, the population of many multi-objective optimizers does not have the anytime property for any Pareto-compliant quality indicator, and only some optimizers are anytime for some quality indicators, such as hypervolume, but not for others~\cite{LiLopYao2023archiving}. The only indicator-independent way to capture the anytime behavior of a multi-objective optimizer is to store all evaluated solutions in an external unbounded archive (or all solutions which are non-dominated at the time of evaluation, if the indicator is Pareto-compliant). 

Detailed surveys on archiving in MOO are available in \cite{LiLopYao2023archiving,SchHer2021archiving}. 
These include bounded archivers, which maintain archives within a fixed size limit, often using adaptive grid selections or hypervolume approximations of the Pareto front. Other strategies use $\epsilon$-dominance to create respective finite-size approximations. However, our approach of using an unbounded external archive~\cite{FieEveSing2003tec,ishibuchi2020new,TanOya2017gecco} allows for maximum flexibility for parsing solutions during the analysis step for further performance analysis in an anytime fashion: 
The value of any Pareto-compliant quality indicator applied to this archive after each solution evaluation is monotonic. Thus, we can employ the same techniques for anytime analysis as in the single-objective case.

\subsection{Algorithm Comparisons / Rankings}

When comparing algorithm performance, we are typically interested in how algorithms perform across a set of problem instances, multiple independent runs, or a combination of both~\cite{MersmannTNW10}. 
Two common approaches are ranking algorithms based on their performance (indicators) on individual instances or by examining the aggregated/mean performance. 
Instance-level ranking provides insights into the relative performance differences but does not account for absolute counterparts. This approach is also used to test for statistical difference, e.g., the Friedman test, and is commonly visualized using Critical Difference plots~\cite{Demsar06}.
Mean performance, on the other hand, does account for these absolute performance differences; however, it is susceptible to permutations in the underlying instance distribution. Furthermore, obtaining statistical insights is difficult without making further assumptions about the underlying distributions.

A recently introduced, more robust alternative to comparing algorithms is to take a bootstrapping approach~\cite{FawEtAl23}. Here, many resamples of the problem instance set are created, after which algorithms are ranked based on their overall performance in each resample. Statistical guarantees can be obtained with the resulting ranks while focusing on overall performance. In another work, this method was extended to obtain robust rankings when multiple performance objectives are considered~\cite{DBLP:conf/gecco/RookHT24}.

\section{IOHprofiler for Multi-Objective Optimization}

\begin{figure}[t]
    \centering
    \includegraphics[width=0.9\linewidth]{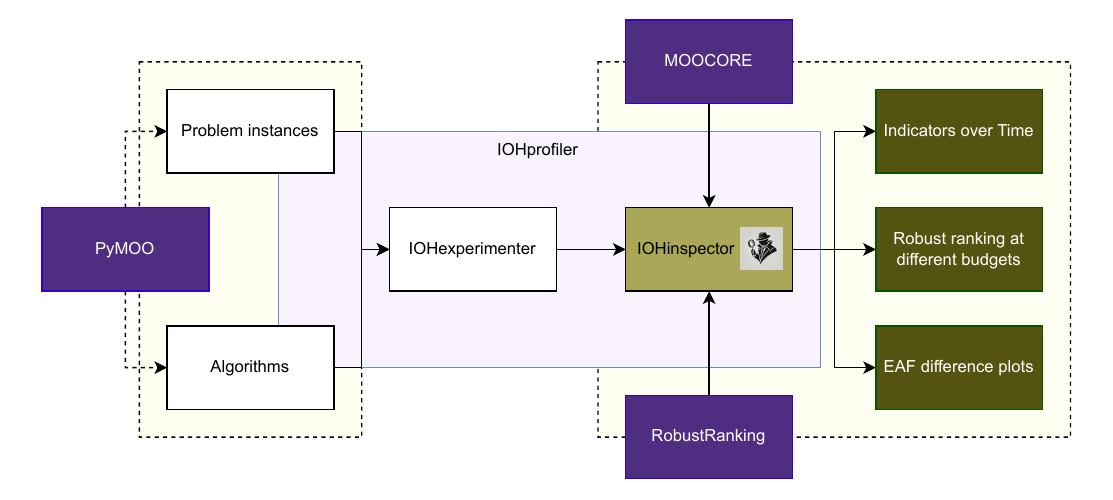}
    \caption{Schematic Overview of our benchmarking pipeline.}
    \label{fig:schematic}
\end{figure}

This work proposes a new module for benchmarking multi-objective optimization algorithms that fits into the existing IOHprofiler framework. IOHprofiler is a modular set of tools that comprises different parts of the benchmarking pipeline: IOHexperimenter~\cite{IOHexperimenter2024} is mainly used for performing experiments and ensuring consistent data logging. In contrast, IOHanalyzer~\cite{IOHanalyzer} can use those logs to perform various analyses and create corresponding visualizations. This separation between experimentation and analysis allows data to be re-used between experiments, even when the analysis has different goals. 

Our proposed extension to IOHprofiler consists of two parts: a preliminary logging methodology (using IOHexperimenter~\cite{IOHexperimenter2024}), which we link with the existing PyMOO library~\cite{DBLP:journals/access/BlankD20} to showcase its functionality, and a Python package for analyzing the resulting performance logs, called `IOHinspector'. Note that this does not integrate with the existing IOHanalyzer website; instead, it serves as a purely local package that supports a subset of visualization methods for single-objective optimization. In the remainder of this section, we will outline the key functionalities and design decisions underlying the multi-objective components of IOHinspector.

To preserve the separation between experimental design and analysis methods, simply logging the value of a chosen indicator would not be sufficient. Instead, we log all (or optionally all non-dominated~\cite{DiasAchiveSoftware}) points evaluated during the search, from which we can then re-compute any indicator value during analysis. This way, we essentially use an unbounded archive strategy (see Section 2.1) for data logging. Since this archive only needs to observe the calls from the algorithm to the problem, the logging can remain independent from the algorithm's implementation, simplifying its use for arbitrary benchmarking setups. Figure~\ref{fig:schematic} outlines the way in which these tools can integrate with a standard benchmarking setup, as we will show in more detail in Section~\ref{sec:exp}.  

Since a benchmarking study can have a wide variety of goals, the flexibility of the analysis pipeline is key. To achieve this, we start from the most informative set of logged data and reduce this to the required indicators or measures depending on the specific analysis or visualization method. In particular, we provide interfaces to various performance indicators, each of which can be calculated at any point in the search. However, since some indicators can be computationally expensive, we perform the calculation lazily, i.e., only at the points used in the resulting visualization (which is user-adjustable). Additionally, all indicators that rely on some problem-specific input, such as reference points or sets, can be modified during the analysis. 

Our current version of `IOHInspector' contains the following indicators: hypervolume~\cite{ZitThi1998ppsn}, IGD+~\cite{DBLP:conf/emo/IshibuchiMTN15}, R2~\cite{DBLP:conf/gecco/BrockhoffWT12}, Epsilon~\cite{ZitThiLauFon2003:tec} 
(additive and multiplicative versions). The majority of these indicators are computed by the \texttt{moocore} package\footnote{\url{https://multi-objective.github.io/moocore/python/}}, which provides efficient implementations for their calculation.

\section{Experiments}\label{sec:exp}

To illustrate the benefits of the anytime approach to benchmarking MOO algorithms and showcase the proposed framework's functionalities, we perform a benchmarking study on a set of well-known problems and MOEAs. To ensure full reproducibility and showcase how to use our proposed toolbox, we provide a full set of reproducibility documents on our Zenodo repository~\cite{reproducibility}. This repository contains several notebooks that outline the steps taken to produce the included results, as well as some additional analyses and visualizations that could not be included in this paper.  The IOHinspector package can also be found on GitHub (\url{https://github.com/IOHprofiler/IOHinspector}).

\subsection{Experimental Setup}

Since we focus on integration with existing tools for MOO, we select two of the most well-known problem suites as our benchmark set: ZDT~\cite{ZitThiDeb2000ec} and DTLZ~\cite{DebThiLau2005dtlz}. For both sets, we utilize the default settings from PyMOO. This means that the problems from ZDT have $d=2$ and those from DTLZ have $d=3$ as their objective space dimensionality, respectively. 

As algorithms we use NSGA-2 \cite{Deb02nsga2}, 
SMS-EMOA \cite{BeuNauEmm2007ejor}, 
NSGA-3~\cite{DebJain2014:nsga3-part1}, MOEA/D \cite{ZhaLi07:moead}, RVEA \cite{CheJinOlhSen2016rvea} and random search as baseline.
For each of these algorithms (except random search), we vary the population size\footnote{Or number of reference vectors. Algorithms where both population size and number of reference vectors can be set always have these parameters equal to each other.} to be in $\{10,25,50,100,250,$ $500\}$. When reference vectors need to be provided, such as in MOEA/D, they are generated based on Riesz's-Energy~\cite{ref_dirs_energy} method integrated into PyMOO.

Each algorithm configuration is run $25$ times on each problem, with a budget of $50\,000$ function evaluations. To enable more straightforward normalization in our analysis, we translate each function so its single-objective minima are 0 (for each objective in the objective space).    

Throughout the remainder of this paper, we will use normalization of the objective values to allow for convenient aggregation across problems. While many normalization approaches exist, we opt to use min-max normalization based on the ideal and worst attained point across all runs. 
We note that the proposed software is fully flexible in this aspect, and other normalization approaches can be used without impacting any other design decisions.

\subsection{Hypervolume over Time}
\begin{figure}[t]
    \centering
    \includegraphics[width=0.48\linewidth]{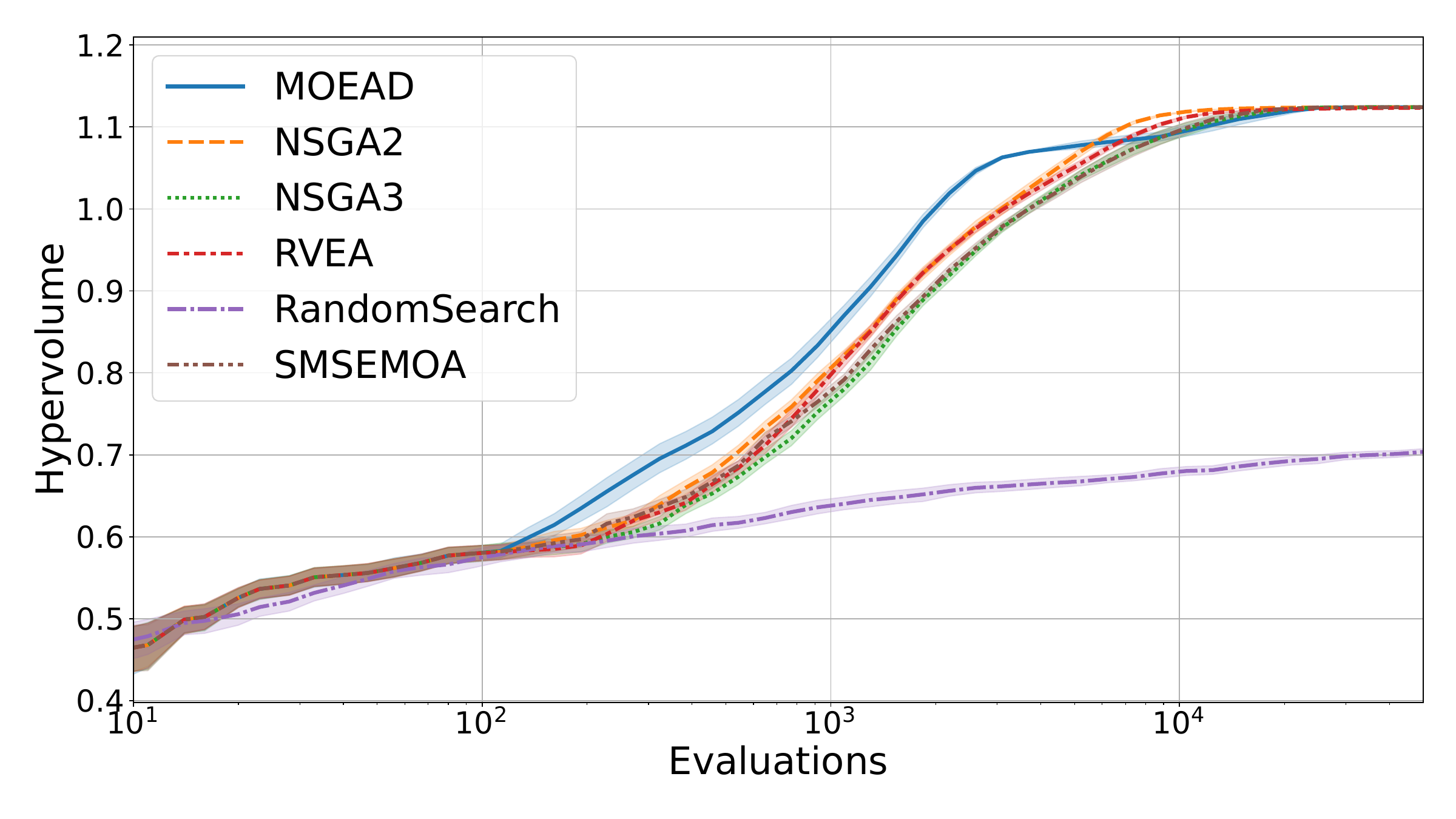}
    \includegraphics[width=0.48\linewidth]{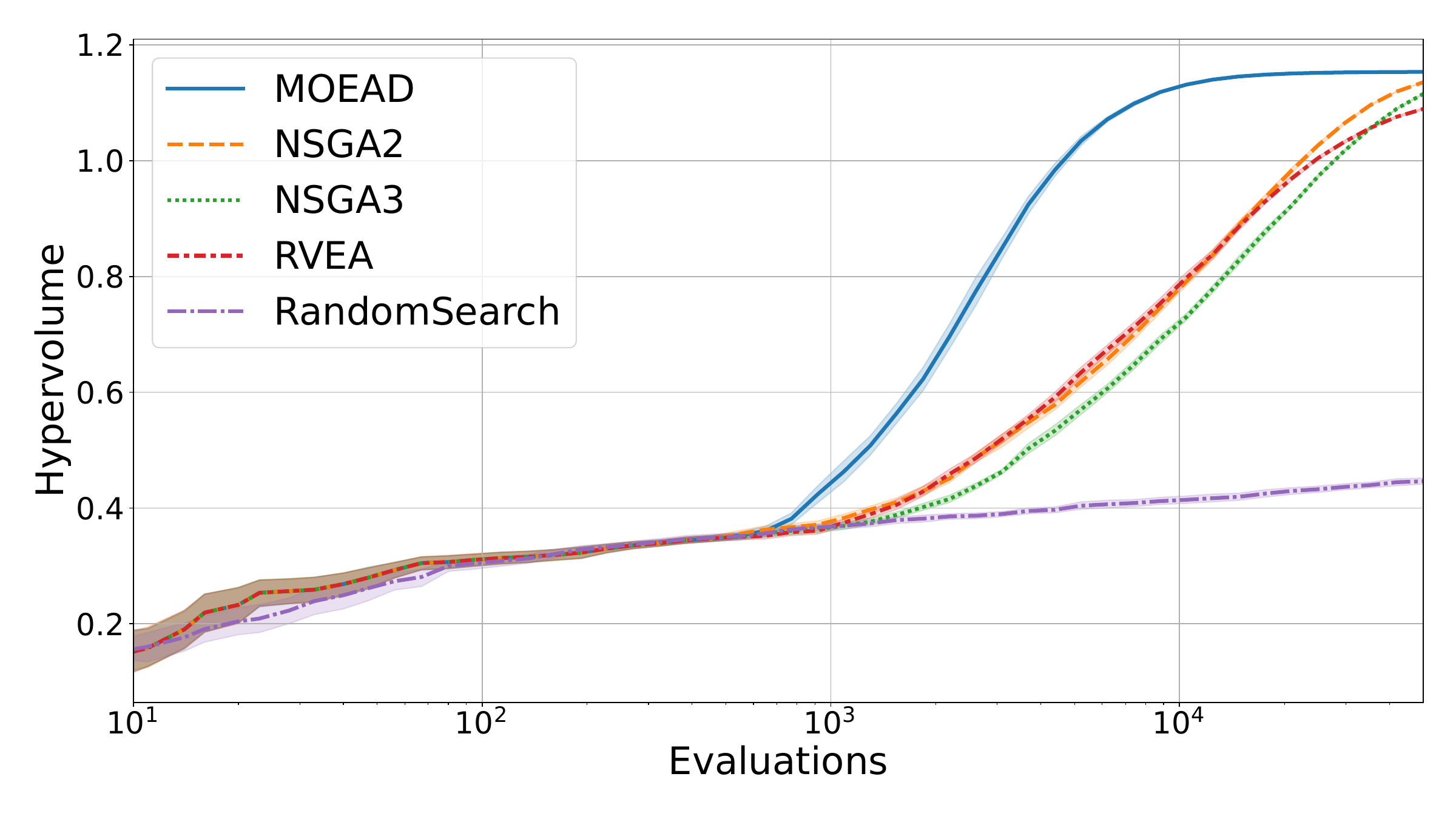} \\
    \includegraphics[width=0.48\linewidth]{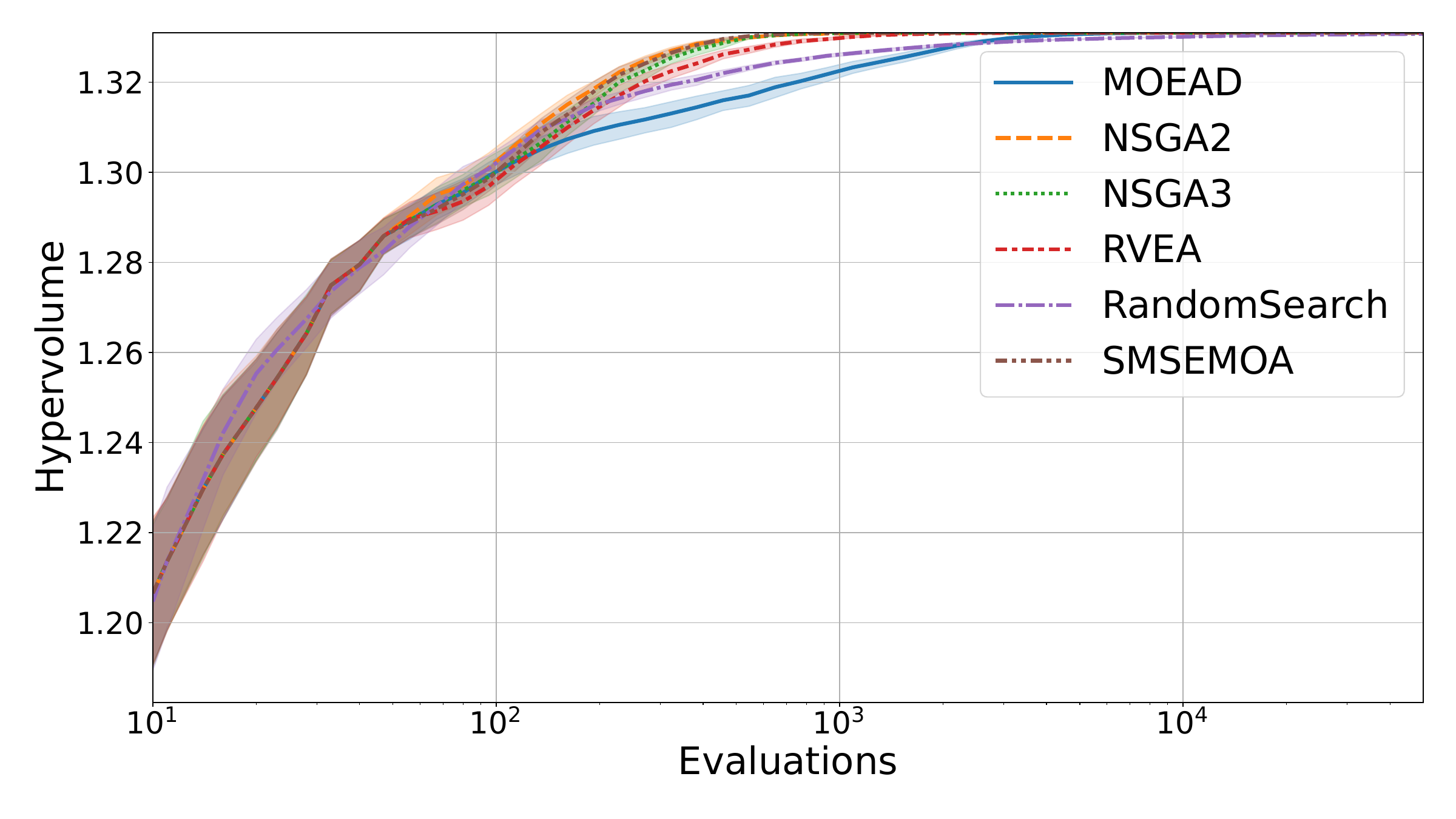}
    \includegraphics[width=0.48\linewidth]{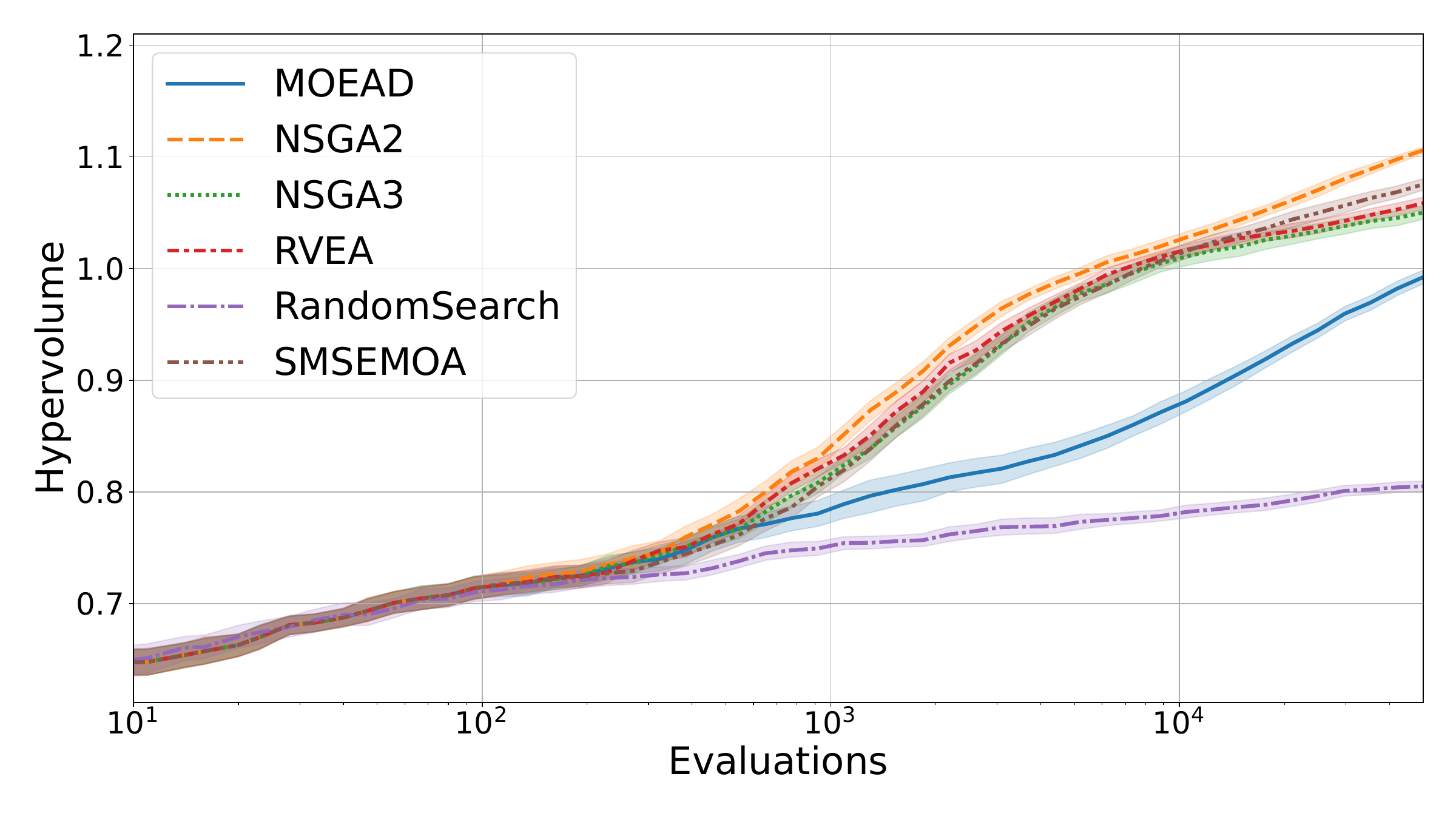}
    \caption{Evolution of hypervolume over time for the selected algorithms on selected problems. Within each subplot, algorithms have been set to a common population size/number of reference vectors. From left to right, top to bottom, the function (population size) of each figure is as follows: ZDT2 (100), ZDT6 (500), DTLZ1 (50), and ZDT5 (100). The reference point is always set to $[1.1]^d$ after normalizing the objectives ($d=2$ for ZDT, $d=3$ for DTLZ). Shaded areas show the 95\% confidence intervals, lines show the mean. }
    \label{fig:example_hv_over_time}
\end{figure}

We focus on the hypervolume indicator for our first set of results. Since we normalized our objective values, we take $[1.1]^d$ as our reference point, following the methodology from~\cite{IshImaSet2018refpoint}. Using this setup, we can compute the hypervolume at any point in the search (every evaluation, not just every generation, as the specific order of evaluation is determined within the algorithm). We visualize this in Figure~\ref{fig:example_hv_over_time}, where we show hypervolume on four selected problems.
We can see, e.g., from the results on ZDT2, that relative rankings between algorithms can vary significantly throughout the search, with MOEA/D starting out very effectively but being overtaken by most other algorithms after $10^4$ evaluations. 
On ZDT6, we see MOEA/D as the best overall algorithm, while the performance of RVEA flattens out, and NSGA-3 surpasses it towards the end. In contrast, on DTLZ1, we can see poor performance of MOEA/D being worse than RandomSearch and only after $10^3$ evaluations catching up with other algorithms and surpassing RandomSearch. On ZDT5, we can see NSGA-2 consistently being the best algorithm and MOEA/D being the worst solver while the other algorithms have similar performance.
From these examples, we see that while, in many cases, multiple algorithms reach similar performance at the end of the search, there are significant differences in the number of evaluations they need to reach this value. Simultaneously, changing rankings for different numbers of evaluations can be observed. 

In addition to the common comparison of different algorithms, we can also compare the performance of different settings of the same algorithm. Here, we focus on population size, as this is usually a parameter that prevents fair comparisons when only the final solution set returned by the algorithm is considered. To illustrate this type of comparison, we show different settings of MOEA/D on ZDT4 on the left side of Figure~\ref{fig:popsize_example}. This figure shows that the hypervolume of the initial random sample is the same until the desired population size is reached. This shows that every population size has the same starting point, and the specific behavior of different population sizes can be analyzed. Most interesting here are the low and high population sizes. Population size 10 has the worst performance with few evaluations but achieves the best performance around $10^3$ evaluations. Population sizes of 250 and 500 have the worst performances but converge after $10^4$ evaluations like the other population sizes.  

\subsection{Changing Indicators}
Since we track all individual objectives rather than a single indicator, we can use the same data for multiple comparisons based on different indicators. This is a clear example of the separation between analysis and experimental setup, which also enables data to be reused for different purposes. In this section, we change the focus to the IGD+ indicator.  

\begin{figure}[t]
    \centering
    \includegraphics[width=0.48\linewidth]{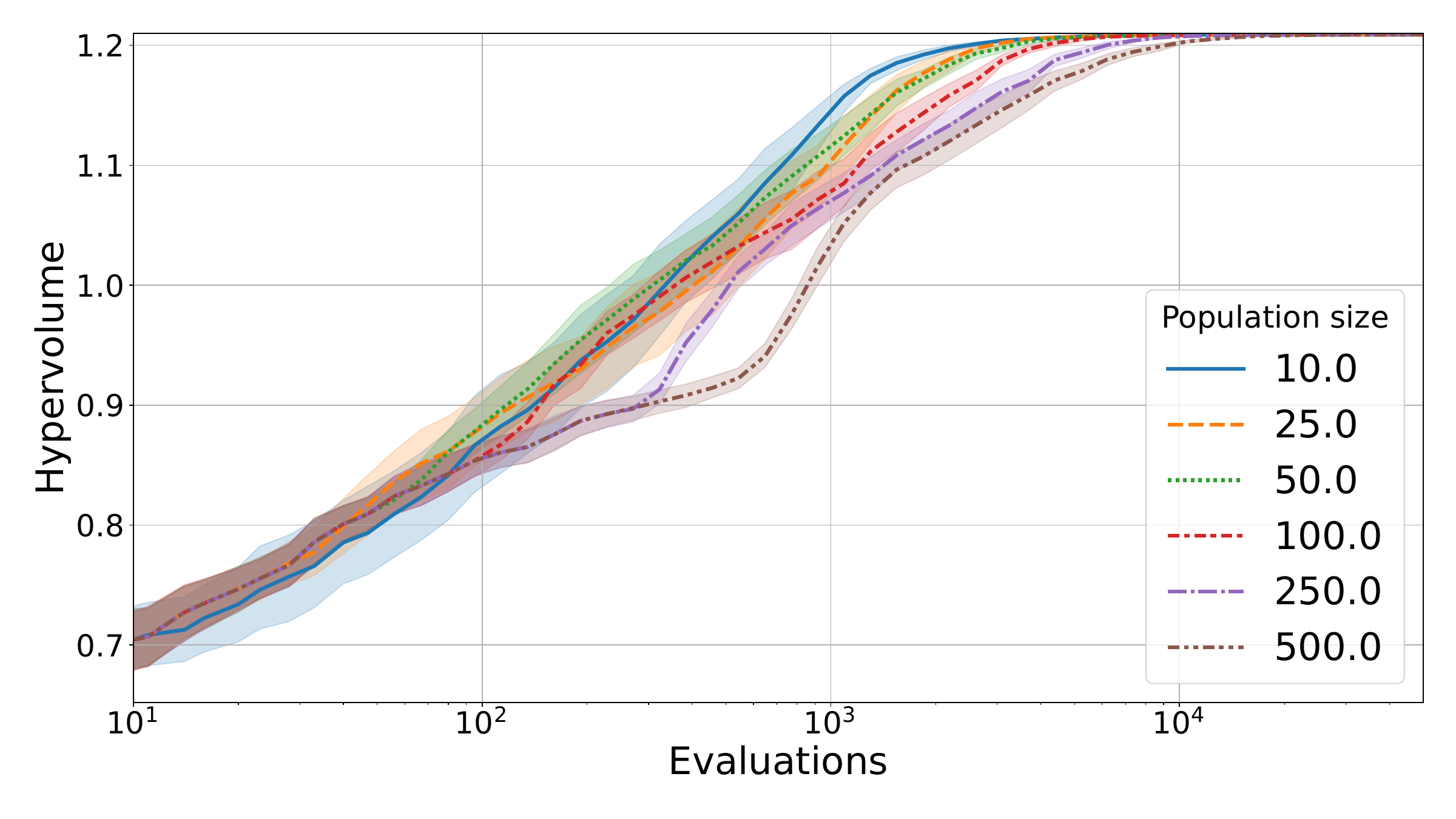}
    \includegraphics[width=0.48\linewidth]{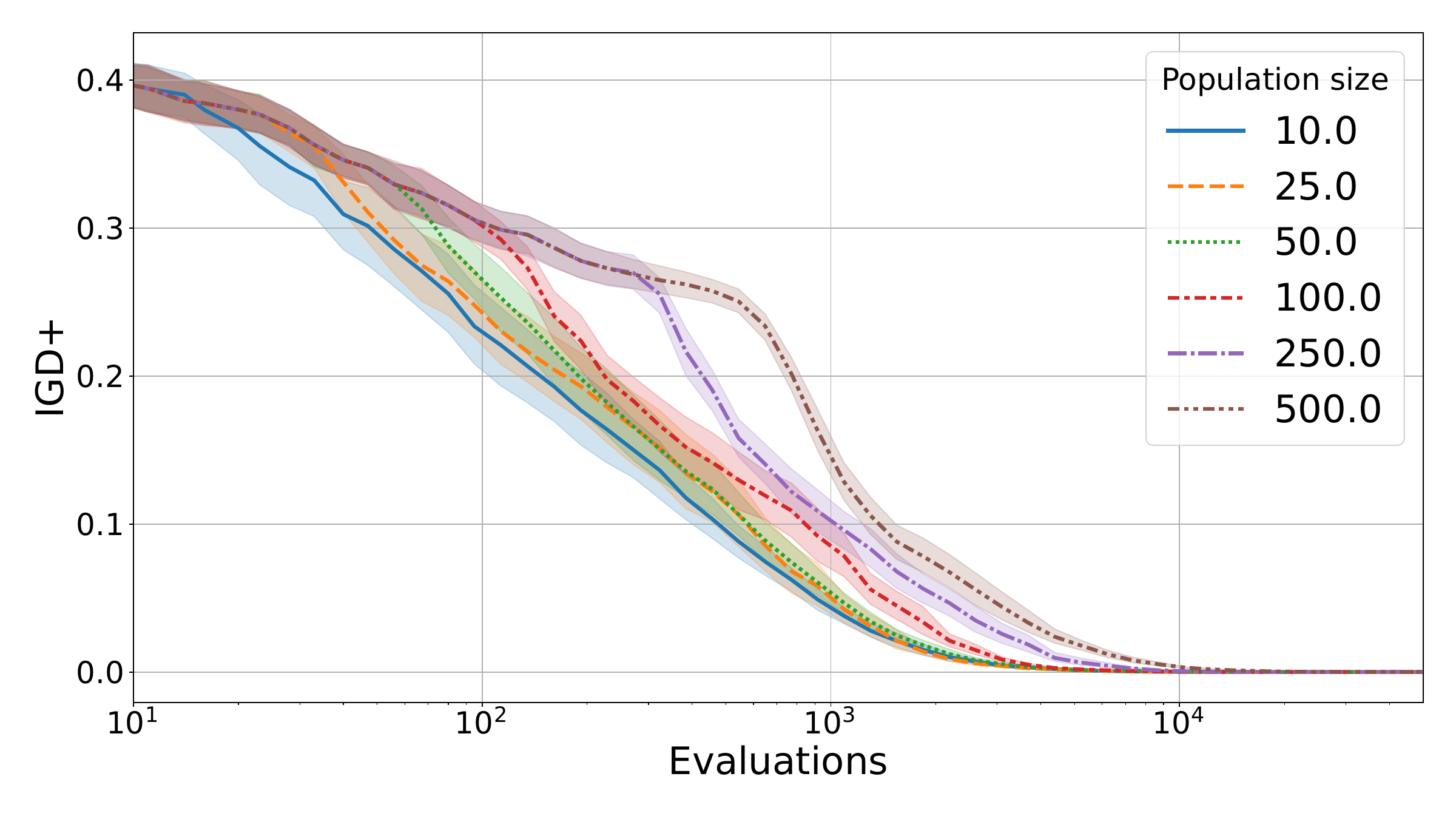}
    \caption{Evolution of hypervolume (left) and IGD+ (right) over time for the selected parameterizations for the MOEA/D algorithm on ZDT4. The used reference set for IGD+ is taken from PyMOO, while the reference point for the hypervolume is set to $[1.1]^2$ after normalizing the objectives. Shaded areas show the 95\% confidence intervals.}
    \label{fig:popsize_example}
\end{figure}

To calculate the IGD+, we need to define a reference set for each problem. In our case, we can either take the provided set from PyMOO or extract a reference set from the existing data. We opt for the former in this paper but provide access in our toolbox to the latter as well (using filtering of data from many runs). 
The resulting performance over time is plotted on the right side of Figure~\ref{fig:popsize_example}, from which we can see that it shows the same characteristics as the hypervolume plot on the left, e.g., the curves for population sizes 250 and 500 have distinctive bends at the same position. These bends visualize the transition from the 1\textsuperscript{st} (randomly initialised) to the 2\textsuperscript{nd} (selected by selection) generation. This visual change is also present for the other poupulation sizes, but to a lesser exend. Population sizes 25, 50, and 100 show a slightly wider spread but converge simultaneously. Noteworthy is the consistently good performance of population size 10.

\subsection{Aggregations: ECDF}

In addition to the per-function evolution of indicator values, we can also take a more aggregated view of our performance data. By normalizing performance to $[0,1]$, we can aggregate, e.g., hypervolume over time for all functions in our benchmark. Note that this normalization is done in addition to the objective-normalization, to allow for aggregating objective spaces of different dimensionality. In our case, this normalization is done by using the maximum hypervolume based only on the reference and ideal points ($[0]^d$ and $[1.1]^d$ respectively), 
but this could also be done, e.g., using the reference front information. These normalized performance values can then be aggregated to provide an overview of the general behavior of the algorithm over the full set of benchmark problems we consider. Using this aggregation of normalized performance is equivalent to the commonly used Empirical Cumulative Distribution Functions, where we don't need to define an a priori set of quality targets\footnote{Instead of targets, we use the normalized performance values directly~\cite{LopVerDreDoe2025}.}.
\begin{figure}[t]
    \centering
    \includegraphics[width=0.9\linewidth]{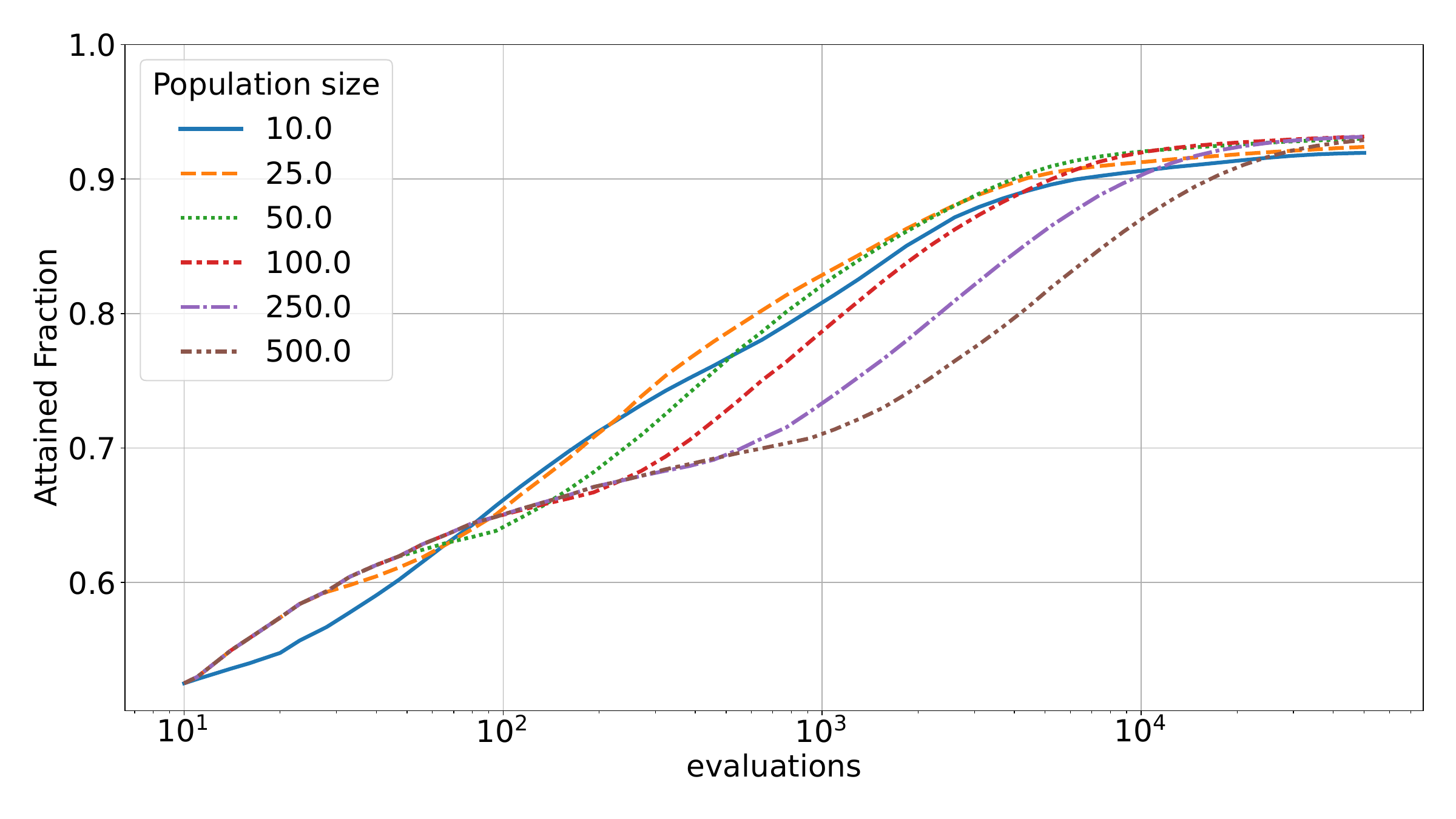}
    \caption{Example plot for ECDF of hypervolume (where the reference point is $[1.1]^d$ and then hypervolume is scaled to $[0,1]$). This plot aggregates the hypervolume over time behavior on all considered benchmark problems for the selected algorithm (NSGA-2) with different population sizes. }
    \label{fig:example_ecdf_hv}
\end{figure}

Figure~\ref{fig:example_ecdf_hv} shows this aggregated EAF-based ECDF \cite{LopVerDreDoe2025} for the hypervolume indicator on NSGA-2. Somewhat surprisingly, we see that, for every population size, the first iteration after initialization generally leads to a worse hypervolume fraction compared to having a larger initial population. We also notice that the algorithm with the highest hypervolume fraction changes over time, with larger population sizes overtaking the smaller ones near the end of the budget. 

\subsection{Comparing Attainment Surfaces}
\begin{figure}[t]
    \centering
    \includegraphics[width=0.45\linewidth]{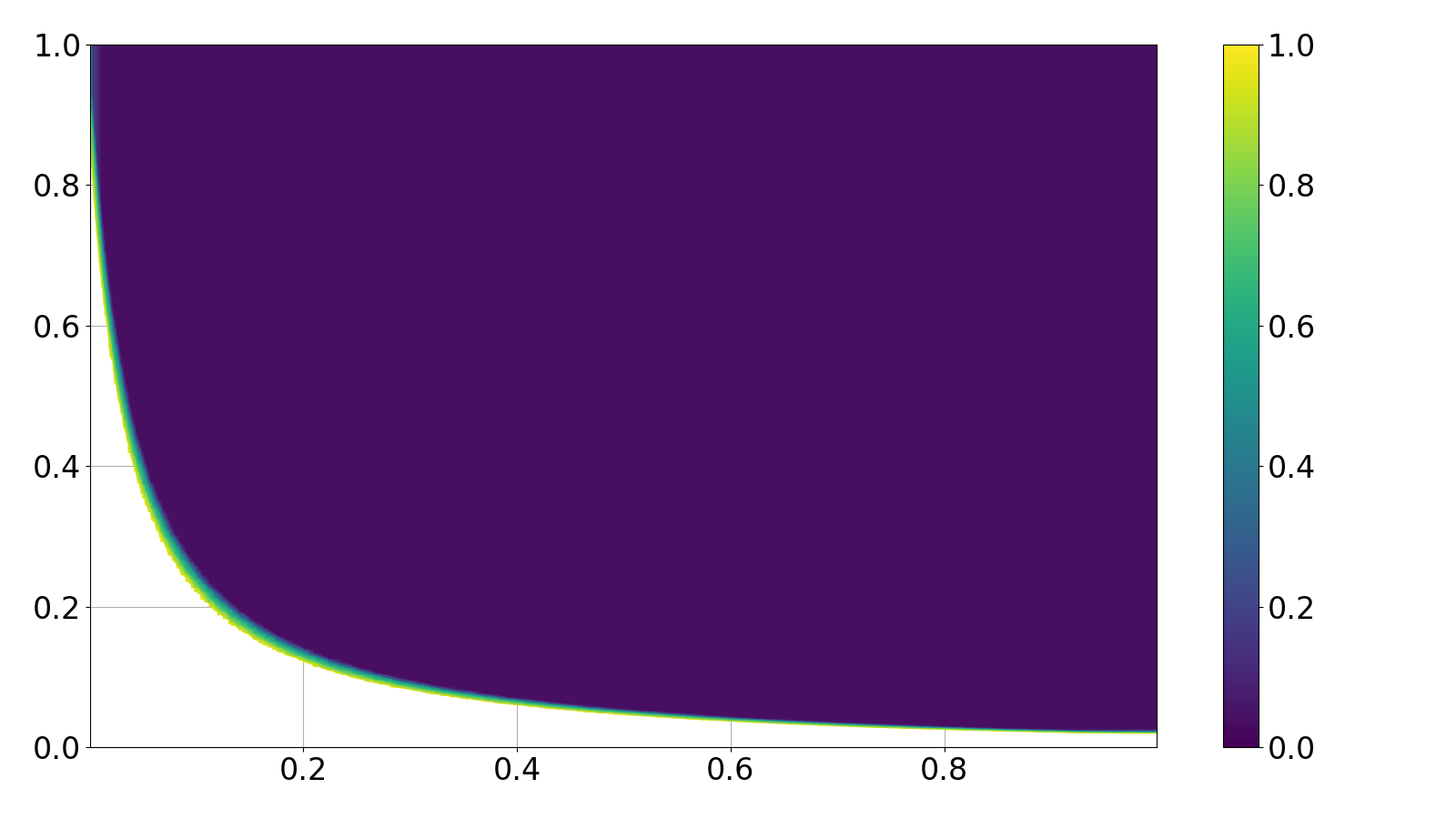}
    \includegraphics[width=0.45\linewidth]{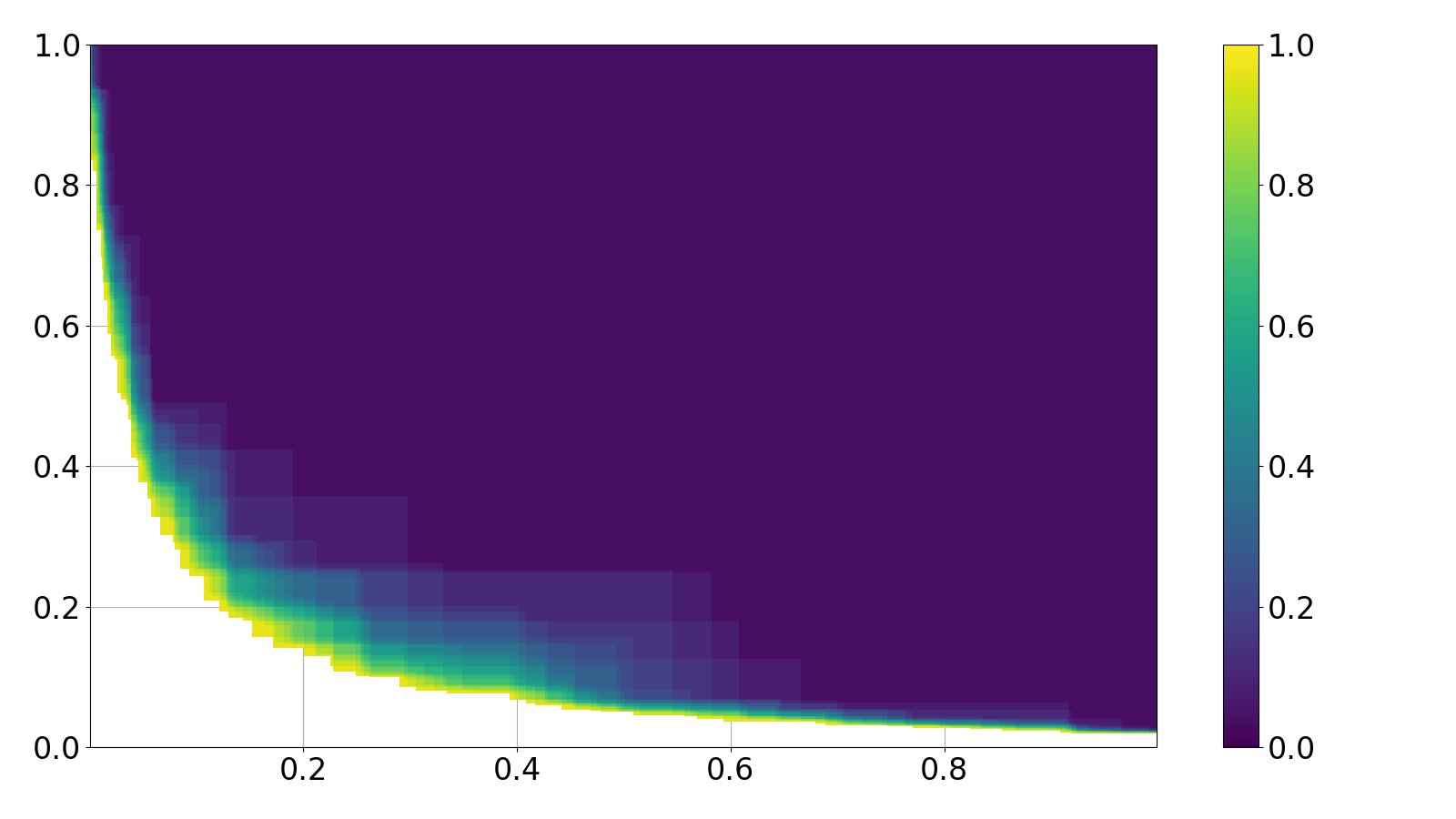}
    \includegraphics[width=0.85\linewidth]{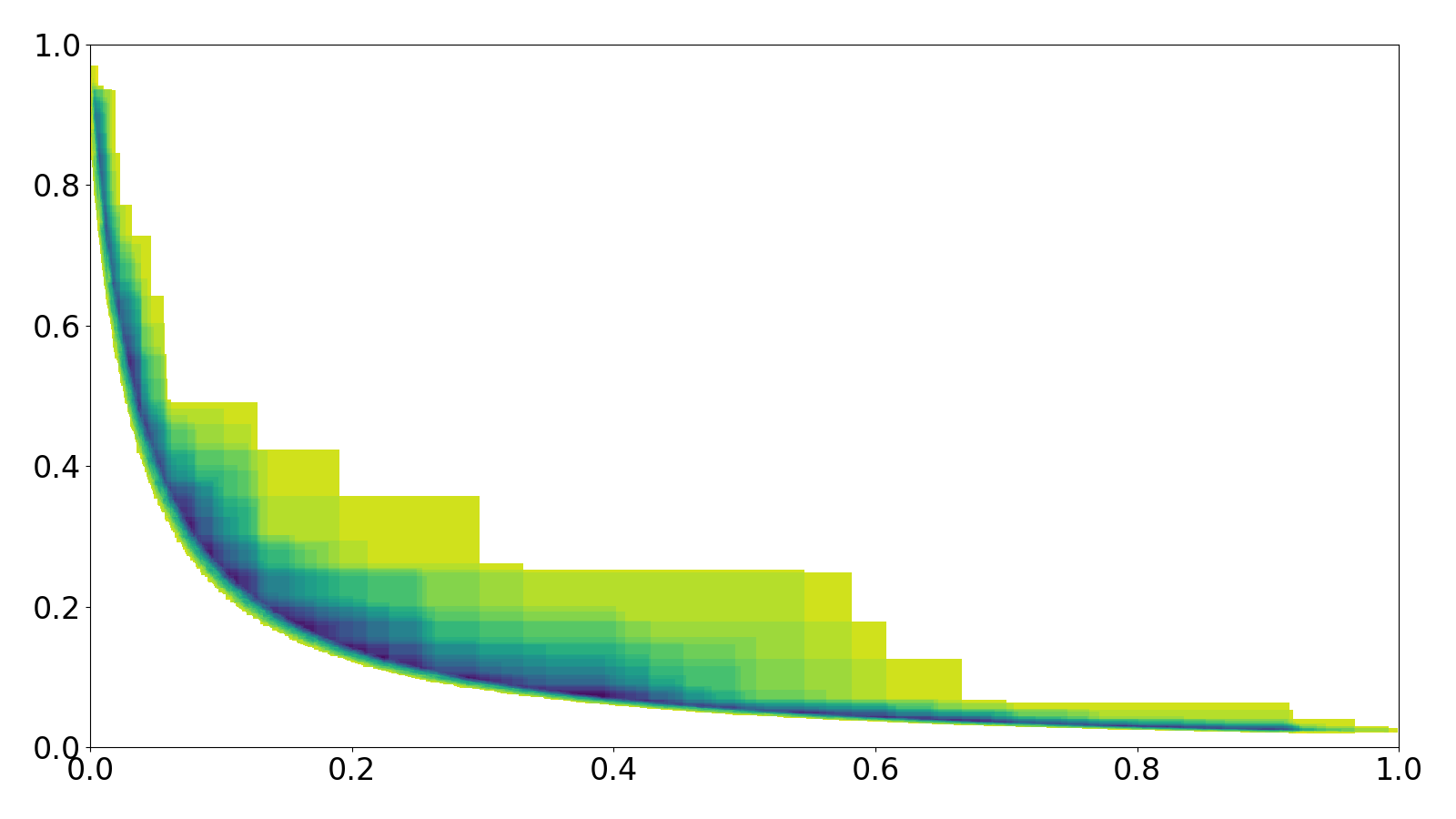}
    \caption{EAF plots of the full set of solutions evaluated by NSGA-2 with population size 10 (top left) and the final set of solutions returned by the algorithm (top right) on ZDT5. The colors indicate the fraction of runs in which a solution dominating the point was attained. The bottom plot shows the EAF-difference between these two plots, where colors correspond to the fraction of runs where a dominating solution was found in the archive, but not in the final population. }
    \label{fig:eafs}
\end{figure}

In addition to the anytime performance, we can also compare the differences in solution sets returned by the algorithm relative to all points it has visited during the search. Corresponding Empirical Attainment Surface (EAF) plots considering the final solution and all points can be seen in Figure \ref{fig:eafs}. Based on these plots, we can create an EAF-difference plot between the returned and full set of solutions (i.e., the whole external archive) to show the regions of the objective space that get `lost' when only looking at the final population. Such an EAF-difference plot for NSGA-2 with population size 10 on ZDT5 is shown in the bottom part of Figure~\ref{fig:eafs}. Here, we can see that the final solution of NSGA-2 is able to mostly cover the outer parts of the possible attainable Pareto front compared to all visited solutions but falls short of consistently covering the middle part.

\subsection{Robust Ranking over Time}
We can also look at other types of comparison at different points during the search. One example is robust ranking~\cite{FawEtAl23}, which makes rankings based on a single indicator at different budgets or even extends to multiple indicators~\cite{DBLP:conf/gecco/RookHT24}.
IOHinspector interfaces with the \texttt{robustranking} package of~\cite{DBLP:conf/gecco/RookHT24} to easily generate different types of rankings, including single- and multi-objective robust rankings. 

\begin{figure}[t]
    \centering
    \includegraphics[width=\linewidth]{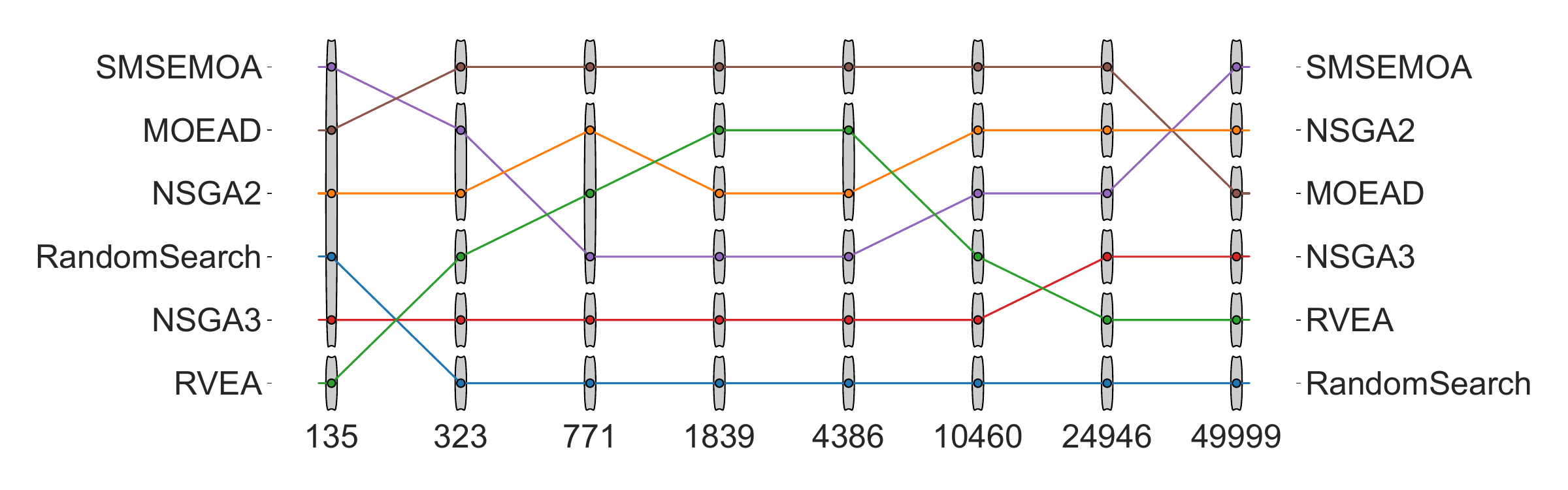}
    \caption{Robust ranking at different evaluation times for the hypervolume indicator on all runs on all problems with population size 100. The budgets are logarithmically selected between $100$ and $50\,000$, and the ranking order starts at the top. The significance threshold $\alpha$ was set for each ranking to $0.05$, and $10\,000$ bootstrap samples were drawn. Statistically tied algorithms are within a gray box.
    }
    \label{fig:robustrank_example}
\end{figure}

\begin{figure}[!t]
    \centering
    \begin{subfigure}[b]{0.48\textwidth}
        \centering
        \includegraphics[width=\linewidth]{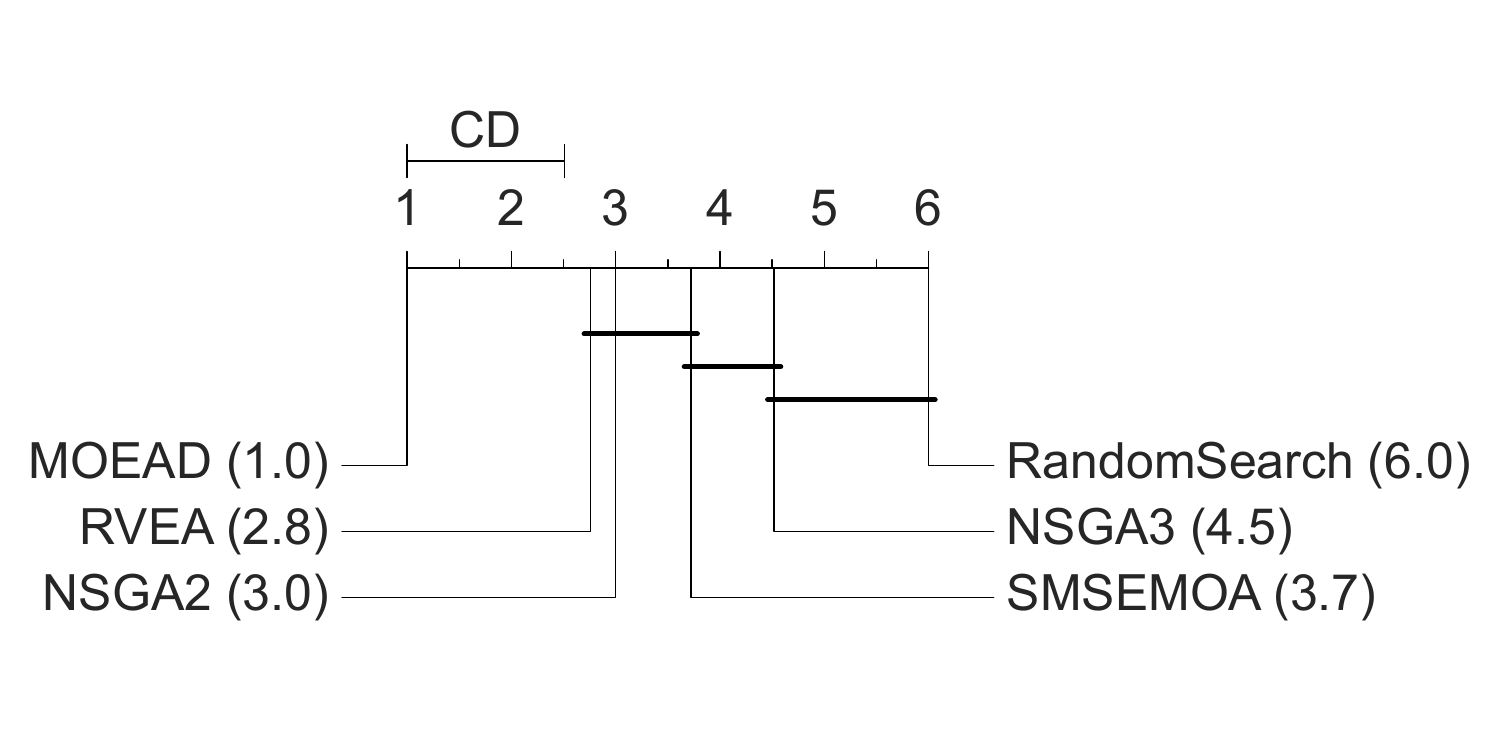}
        \caption{Critical difference plot after $4\,386$ evaluations. $\alpha=0.05$.}
        \label{fig:cdplot}
    \end{subfigure}
    \hfill
    \begin{subfigure}[b]{0.48\textwidth}
        \centering
        \includegraphics[width=\linewidth]{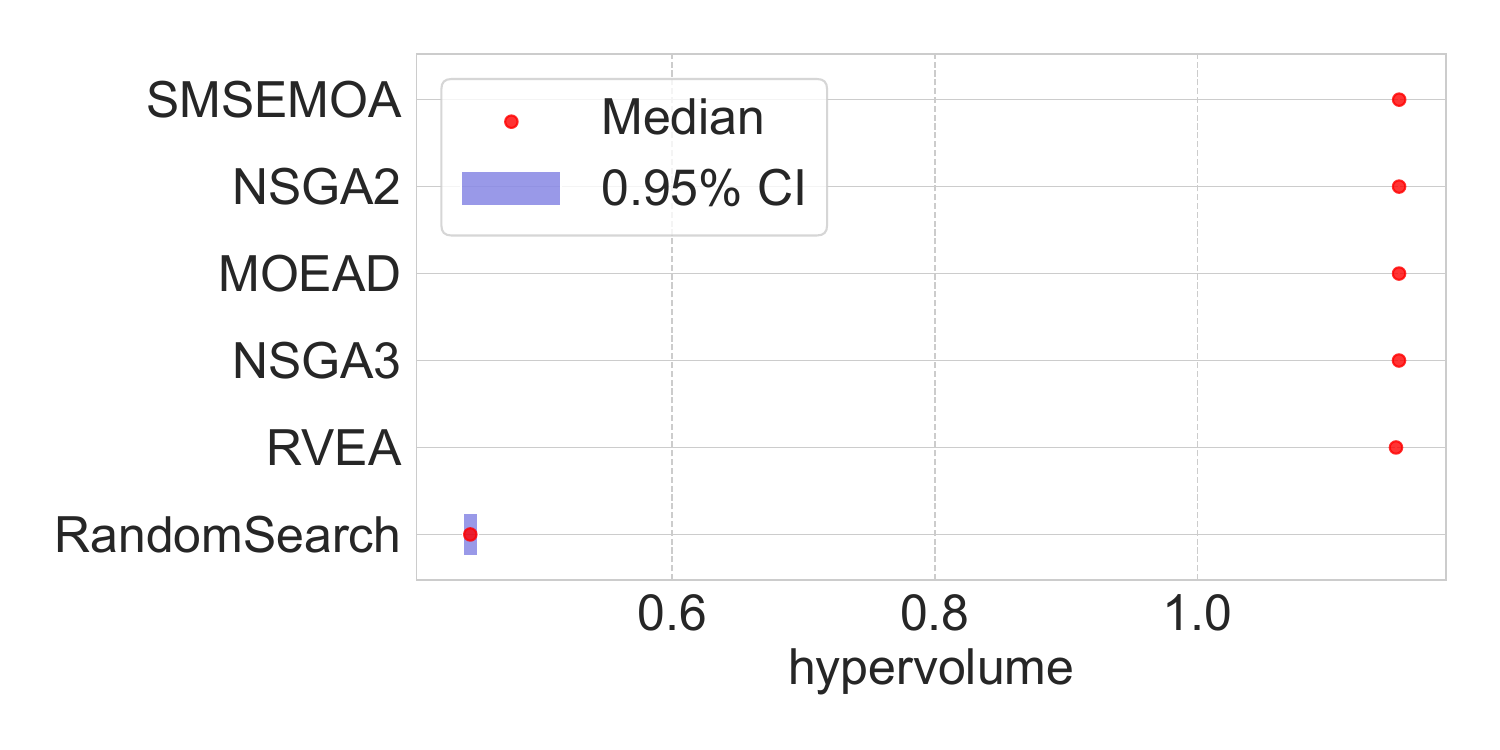}
        \caption{Confidence intervals after $49\,999$ evaluations.}
        \label{fig:ciplot}
    \end{subfigure}
    \caption{Alternative ranking plots originating from the robustranking package.}
\end{figure}
Figure~\ref{fig:robustrank_example} shows the robust ranking of the algorithms with population size $100$ at eight evaluation budgets. The figure shows that the rankings over time are not fixed and provide interesting insights that extend and confirm the other presented visualizations like the ECDFs.

Recall that robust ranking assesses (mean) overall performance. The rankings are different compared to other commonly used ranking approaches, such as critical difference plots~\cite{Demsar06}. Figure~\ref{fig:cdplot} shows the CD plot after $4\,386$ evaluations. Although the ranking order -- in this example -- is equal, the statistical ties are not. This is because both rankings compare different aspects: overall performance versus individual performance. 

In the last ranking in Figure~\ref{fig:robustrank_example}, SMS-EMOA surpasses MOEA/D, which was in the first group in all earlier rankings. However, the confidence intervals of the bootstrap distribution in Figure~\ref{fig:ciplot} show that the confidence intervals are very narrow, and the median values of the MOEAs are close to each other. This suggests that these algorithms, in their underlying samples, all tend to have converged.  
In turn, this indicates that the used problem instances are not extremely challenging enough for these algorithms. 

As with all visualization and analytic perspectives provided in this paper, the settings of IOHinspector can be modified based on the specific goals of a user's benchmarking setup. This enables a wide range of different viewpoints on the algorithmic performances and the respective strengths and weaknesses of different algorithms in the considered portfolio.

\section{Conclusions and Future Work}

In this paper, we explored several aspects of benchmarking multi-objective optimization algorithms. Expanding on the existing IOHprofiler framework, we proposed software tools to simplify and support the anytime-performance comparison of multi-objective optimizers. By utilizing unbounded archives for data logging, we showcase that this anytime analysis can be achieved without modifying existing algorithms. Following the design principles of IOHprofiler, we aimed to separate decisions on the analysis side of the benchmarking pipeline from the experimental design choices during the data collection. This way, performance data can be used and re-used to answer various research questions. For example, this approach allows for changing indicators, reference points, and even ranking methodology based on the goals of the analysis.

The software presented in this work consists of two parts. The first is an update to the IOHexperimenter to enable logging multiple objectives using an unbounded archive. The second is the new IOHinspector package, which integrates a variety of multi-objective specific analysis methodologies, such as different indicators and ranking schemes, with visualizations commonly used in single-objective optimizations. As such, our IOHinspector package is not specific to the multi-objective case and can be used more broadly. To illustrate the functionality of these IOHprofiler modules, we integrated them with the popular PyMOO package. In the future, we also aim for more direct integration into other popular libraries for multi-objective optimization. In particular, integrating subset selection mechanisms to extract finite-size approximations of the Pareto front from the stored archive would be very beneficial.

For the analysis of benchmark data, we hope to encourage wider sharing of (reproducible) performance data, which could then be analyzed from different perspectives, e.g., by utilizing different indicators. While we support several commonly used indicators, we aim to expand this selection over time, specifically concerning multi-modal MOO. Finally, we hope these tools can become integrated into other types of benchmarking pipelines, such as algorithm selection and configuration scenarios, to enable the exploration of a wide variety of algorithmic design questions. 

\subsubsection{Acknowledgements} The authors from Paderborn University, University of Twente and Leiden University acknowledge support from the European Research Center for Information Systems (ERCIS). This work was also supported by CNRS Sciences informatiques via the AAP project IOHprofiler, and in part by the COST action CA22137 (ROARNET).

\bibliographystyle{splncs04}
\bibliography{bib/abbrev,bib/journals,bib/authors,bib/articles,bib/biblio,bib/crossref,bibliography}

\end{document}